\documentclass[10pt, a4paper]{article}
\usepackage{lrec}
\usepackage{multibib}
\newcites{languageresource}{Language Resources}
\usepackage{graphicx}
\usepackage{tabularx}
\usepackage{soul}

\usepackage{epstopdf}
\usepackage[utf8]{inputenc}

\usepackage{hyperref}
\usepackage{xstring}

\usepackage{makecell}

\usepackage{color}

\usepackage{graphicx}
\usepackage{float}
\usepackage{listings}
\usepackage{subcaption}
\usepackage{multirow}
\usepackage{tikz}
\usetikzlibrary{decorations.pathreplacing,angles,quotes}
\usepackage{amsmath}
\DeclareMathOperator*{\argmax}{\arg\!\max}

\usepackage{algorithm, algpseudocode}
\usepackage{array}
\usepackage{tabu}
\usepackage{tabularx,booktabs} 
\usepackage{adjustbox}

\makeatletter
\usepackage{microtype}
\makeatother

\usepackage{hyperref}

\newcommand\crule[3][black]{\textcolor{#1}{\rule{#2}{#3}}}

\usepackage[pscoord]{eso-pic}

\newcommand{\placetextbox}[3]{
  \setbox0=\hbox{#3}
  \AddToShipoutPictureFG*{
    \put(\LenToUnit{#1\paperwidth},\LenToUnit{#2\paperheight}){\vtop{{\null}\makebox[0pt][c]{#3}}}%
  }%
}%

\title{Towards Large-Scale Data Mining\\for Data-Driven Analysis of Sign Languages}

\name{Boris Mocialov$^1$, Graham Turner$^2$, Helen Hastie$^3$}

\address{$^1$School of Engineering and Physical Sciences, Heriot-Watt University, Edinburgh, UK\\$^2$School of Social Sciences, Heriot-Watt University, Edinburgh, UK\\$^3$School of Mathematical and Computer Sciences, Heriot-Watt University, Edinburgh, UK \\
         \{bm4, g.h.turner, h.hastie\}@hw.ac.uk\\}

\abstract{
Access to sign language data is far from adequate. We show that it is possible to collect the data from social networking services such as TikTok, Instagram, and YouTube by applying data filtering to enforce quality standards and by discovering patterns in the filtered data, making it easier to analyse and model. Using our data collection pipeline, we collect and examine the interpretation of songs in both the American Sign Language (ASL) and the Brazilian Sign Language (Libras). We explore their differences and similarities by looking at the co-dependence of the orientation and location phonological parameters. \\ \newline \Keywords{Sign Language Data Mining, Phonological Parameter Co-Dependence} }

\begin{document}

\maketitleabstract

\section{Introduction}
The data-driven field of automated sign language understanding is dependent on large amounts of high-quality data, independent of the application or the motivation of the research. Unfortunately, such data have typically had restricted access, either due to the projects finishing or limiting license terms. Online services, on the other hand, offer large amounts of data that have more relaxed terms and conditions and are available for as long as the service providers exist. The field of written and spoken languages has recently greatly benefited from the use of such data for natural language understanding projects trained on, for example, Reddit, Twitter, Amazon reviews.

However, there is an uneven distribution of users of spoken languages versus sign language users around the world. Therefore, the amount of sign language data on the internet is naturally lower than that of spoken languages. Moreover, sign languages do not have a common writing system as opposed to spoken languages, which makes it very difficult to annotate. Furthermore, the data from research projects usually have application-dependent annotation using either words in written languages, phonological parameters, or sign pictures \cite{reinerkonrad2015}. Despite the fact that there is no common writing system for sign languages, HamNoSys defines one notation system that is often used by researchers. This system distinguishes phonological parameters (e.g. location, orientation, movement, handshape, non-manual gestures) present during signing \cite{hanke2004hamnosys}.


First, we show that it is possible to utilise social networking platforms to support research in data-driven automated sign language understanding. Second, we take a look at two sign languages that have relatively little historical relationship. One sign language being ASL and the second being Libras and investigate the signing behaviour of the spoken song interpreters, while looking at three English songs: `Love Yourself' by Justin Bieber, `Halo' by Beyoncé, and `Love On The Brain' by Rihanna. We investigate and compare two phonological parameters: hand location relative to the signers' body and extended finger orientation. This work will quantify frequently occurring hand positioning during the signing and compare the prevailing hand positions and orientations between the two sign languages, aiming to show that sign languages evolve differently. The reason why we investigate interpreted songs is because we want to compare sign languages by looking at continuous signing in different sign languages that sign the same information. Findings in this paper could also assist researchers who work on developing models for sign language understanding by reducing the search space of the models during the optimisation by ignoring combinations that are relatively infrequent during continuous signing.

\section{Motivation for Automated Processing of Sign Languages}
With the rise of accurate pose prediction and hand estimation libraries such as the OpenPose \cite{cao2018openpose,simon2017hand,cao2017realtime,wei2016cpm}, researchers in the field of automated sign language understanding are now able to focus on high level abstract research ideas. Contemporary research looks at translating written languages to sign languages and vice versa, thus resembling research done in the field of the machine translation for spoken languages \cite{cihan2018neural,stoll2019text2sign,yuan2019large}. 

The common linguistically inspired approach is for the raw visual modality of the sign languages to be broken down into the sub-lexical phonological parameters (e.g. location, orientation, movement, handshape, and non-manual gestures). Multiple previous works have modelled individual phonological parameters. \newcite{cooper2007large} modelled hand location, movement, and relative hand position and called them the sub-sign units. \newcite{cooper2012sign} relied on handshape, location, movement, and relative hand position in their work on recognition of the individual signs in The British Sign Language. \newcite{buehler2009learning} examined movement, handshape, and orientation while matching the combination of these parameters to find similar signs. Also \newcite{buehler2010employing} used location and handshape in the multiple instance learning problem. \newcite{7780781} focused on modelling sixty handshapes. Their model is a chain of convolutional neural networks (VGG) pre-trained on the ImageNet data \cite{simonyan2014deep}. In our work, we generate linguistic annotations in the form of hand location relative to the signers' body and extended finger orientation for the continuous interpretations of the three English songs (mentioned above).

The lack of the text annotation that could provide context remains an issue for video data. \newcite{joze2018ms} noticed that many signing videos have captions, which could be an additional source of annotation, as more and more content is being generated online, including that for the deaf community. In this work, we focus only on the linguistic annotations without inferring the context.

\bgroup
\def\arraystretch{1.5}
\begin{table}[H]
    \centering
    \resizebox{\columnwidth}{!}{%
    \begin{tabular}{|l|l|}
    \hline
        \makecell[c]{\textbf{Type}} & \makecell[c]{\textbf{Motivation}} \\ \hline
    
        \multirow{3}{*}{Recognition} & Context-Specific \cite{ko2019neural}  \\
        & Isolated Signs \cite{zhou2009adaptive} \\
        & Individual Phonological Parameters \cite{cooper2012sign} \\ \hline
        
        \multirow{2}{*}{Translation} & Sign-Text \cite{cihan2018neural} \\
        & Text-Sign \cite{stoll2019text2sign} \\ \hline
        
        \multirow{3}{*}{Learning} & Zero-Shot \cite{bilge2019zero} \\
        & Clustering \cite{10.1007/978-3-642-12214-9_18} \\
        & Augmentation \cite{Mocialov2017TowardsCS}\\ \hline
        
        \multirow{2}{*}{\shortstack[l]{Linguistic\\Studies}} & \multirow{2}{*}{Phonological Parameter Co-Dependence \cite{ostling2018visual}} \\ 
        & \\ \hline
        
        \multirow{2}{*}{Education} & Teaching \cite{stefanov2017real} \\
        & Edutainment \cite{5771325} \\ \hline
        
        \multirow{2}{*}{\shortstack[l]{Sign\\Spotting}} & \multirow{2}{*}{\shortstack[l]{Queries \cite{belissen2019sign}}}\\
        & \\ \hline
    \end{tabular}
    }
    \caption{Research directions in the field of the automated sign language understanding}
    \label{tab:categories}
\end{table}
\egroup

We group surveyed papers by their motivation, omitting works that use data other than a single RGB camera and papers that focus on pose estimation, tracking, or finger spelling, as these do not directly align with research in sign language understanding. Table \ref{tab:categories} categorises the research directions in the field of automated sign language understanding. It can be seen that there are projects that focus on more abstract concepts than learning the isolated signs, such as automated data-driven sign language translation. Apart from the recognition of the signs as an attempt to bridge the gap between the hearing and the deaf communities, assistive tools for digital sign language content annotation are gaining interest. For example, \newcite{takayama2018sign} automatically annotate datasets and \newcite{belissen2019sign} query databases with videos of signs, which could be beneficial for accelerating research in linguistic aspects of sign language.

\section{Methodology}
\begin{figure}[H]
    \centering
    \includegraphics[width=0.59\textwidth, clip,trim={2.5cm 23cm 3cm 4cm}]{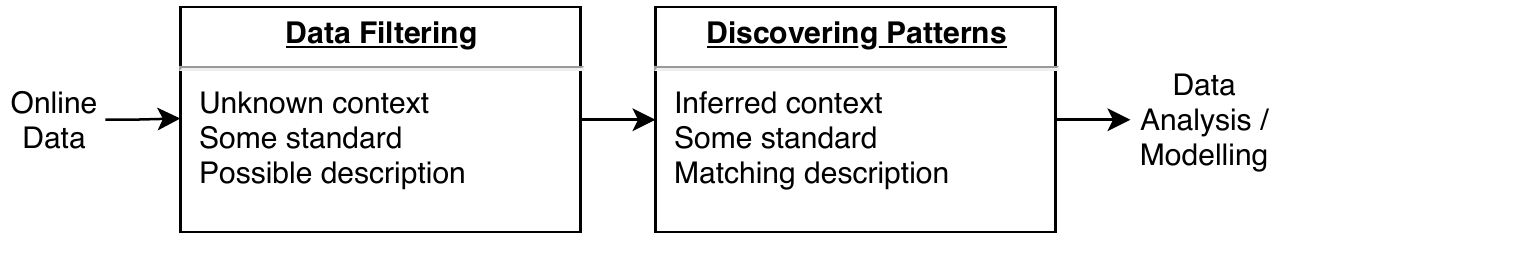}
    \caption{Data collection pipeline for collecting the signing data from the social networking services for the purpose of data-driven automated sign language understanding}
    \label{fig:pipeline}
\end{figure}
Data from online social-media resources tends to be very unpredictable. Therefore, the collection of data has to pass through a number of stages as we suggest in Figure \ref{fig:pipeline}. The first stage of the data collection pipeline is data filtering, where we turn the online data that has no standard into data that has some pre-defined standard. The second stage looks for the patterns in the filtered data either with the help of metadata or the automatic visual analysis of the collected filtered data.

\subsection{Online Data}
In this research, we focus on interpreted songs. As deaf-specific music performers are rare, sign language users resort to `listening' to interpretations of the songs found in the spoken or written languages that are being interpreted by those who can both hear and sign. This is evidenced by the relatively large amount of content found in online resources such as TikTok, YouTube, or Instagram. Such content makes the interpretation of the spoken songs possible for the deaf community, encouraging visualisation of music \cite{Desblache2019}. We consider interpreted songs as our data format because it is possible to find the same songs interpreted in different sign languages, which makes the comparison of the sign languages more precise. We collect one video for every interpreted song for each sign language. Therefore, we have collected a small dataset of continuous signing videos from YouTube from six different signers, interpreting three songs in two sign languages for this proof of concept study.

\subsection{Data Filtering}
We use the OpenPose library for data filtering. The library detects 2D or 3D anatomical key-points, associated with the human body, hands, face, and feet in a single image. The library provides $21$ (x,y) key-points for every part of the hand, $25$ key-points for the whole body skeleton, and $70$ key-points for the face. The OpenPose library helps us apply simple filters to the raw data, discarding all the content that has more than one signer at the same time or any heavy obstructions or occlusions. We also discard the content that has too few key-points visible, as we think it is essential to see the upper body and the hands to make sense of the signing. By performing such filtering, we enforce a quality standard upon the collected online data. However, the context and the signer profile remain unknown.

Other filters could include normalisation, transformation, and rotation of key-points to make the signer appear the same size across videos and to make signers face the camera for more accurate modelling.

\subsection{Discovering Patterns}
We perform visual analysis by extracting the location and orientation sub-lexical phonological parameters from the filtered data looking at the frequency of occurrences of specific location/orientation combinations in the collected filtered data. The following sections will show how we infer sub-lexical components by making use of identified key-points and geometry.

Likewise, metadata can assist in discovering patterns. This metadata can comprise of hashtags, textual description, or the embedded captions on the videos.

\subsubsection{Extended Finger Orientation\\}
A total of eight orientations have been used for the extended finger orientation as defined in the HamNoSys notation with each orientation having $45^{\circ}$ movement (north, north-east, east, south-east, south, south-west, west, and north-west). HamNoSys does define more orientations (e.g. towards or away from the body), however having 2D data makes it difficult to estimate additional orientations. 

The angle is calculated using the inverse trigonometric function between the radius and middle finger coordinates as follows:
\vspace{-1em}
\begin{gather*}
-\pi/2 < \arctan{(q_y - p_y, q_x - p_x)} < \pi/2, \\
\text{where $q$ and $p$ are the $(x,y)$ coordinates} \\
\text{of radius and middle finger metacarpal bones} \\
\text{with every orientation having $\pi/4$ freedom}
\end{gather*}


\subsubsection{Hand Location Relative to the Body\\}
A total of six locations around the body have been used to determine hand position (ears, eyes, nose, neck, shoulder, and abdomen), as opposed to the forty six defined by the HamNoSys notation system. Six were chosen to simplify the detection while complying with the OpenPose library standards. Hand centroids are calculated as follows:
\vspace{-0.8em}
\begin{gather*}
centroid_{right} = (\sum_{i=1}^{N} x_{i_{right}} / N, \sum_{i=1}^{N} y_{i_{right}} / N) \\
centroid_{left} = (\sum_{i=1}^{N} x_{i_{left}} / N, \sum_{i=1}^{N} y_{i_{left}} / N) 
\end{gather*}
where $N$ is the number of points provided by the OpenPose library for each hand.

In order to assign the relative hand location, a threshold has to be assigned as to how far the centroid of a hand can be from a specific body location so as to still be relatively close to that body part. All the distances are measured in pixels and the threshold is set to be $10\%$ of the diagonal of the image frame, which is approximately $100$ pixels. If the distance of a centroid away from all the body parts exceeds the threshold, the hand is considered to be in the `neutral signing space'.

The distance matrix $D$ for every hand is calculated as follows:
\vspace{-0.8em}
\begin{gather*}
M_r \dots N_r = \lvert q_{m \dots n}- centroid_{right} \rvert \\
M_l \dots N_l = \lvert q_{m \dots n} - centroid_{left} \rvert \\
D = 
 \begin{pmatrix}
  M_r & M_l \\
  \vdots  & \vdots  \\
  N_r & N_l 
 \end{pmatrix}
 \end{gather*}

Where $q_{m \dots n}$ are the $(x,y)$ position of the body parts, defined by the OpenPose library (e.g. nose, neck, shoulder, elbow, etc.) and the $M_r \dots N_r$ and $M_l \dots N_l$ are the Euclidean distances between the body parts and right and left hand centroids. 

In order to find the body part $B_{right}$ or $B_{left}$, which has the smallest distance to the centroid of the right or left hand, we use 
\[B_{right} = \argmax D_{i,1}\] 
\[B_{left} = \argmax D_{i,2}\]
The distances are then compared to a threshold to determine if a hand is near a particular body part or is in the `neutral signing space' anywhere around the body.

\subsection{Data Analysis / Modelling}
Once the data has been filtered and the patterns have been discovered, we acquired information on $43016$ hand locations and the same number for the hand orientations for ASL and $38258$ for both hand location and orientation for Libras for the interpreted three songs. We are interested in the analysis of the co-dependence of phonological parameters for each hand and comparing the significant co-dependences across the two sign languages.

\subsubsection{Phonological Parameter Co-Dependence}
Here, we refer to location as $TAB$ and to orientation as $ORI$ for the shorthand notation. First, a global $C_{ORI_N,TAB_M}$ contingency table is generated and counts the occurences for both location/orientation variables for every category that occurs in the collected data (e.g. North, North-East, etc. for orientation and Shoulder, Neck, etc. for location). Second, a series of local contingency tables $C_{2\times 2}$ are constructed from the global $C_{ORI_N,TAB_M}$ contingency table for every category of every variable as a post-hoc step. Finally, Bonferroni-adjusted $p$-value was used \cite{Bland170} to check if the presence of a particular location/orientation combination in the data set is significant, compared to other location/orientation combinations, by performing a Chi-square test of independence of variables for all the $C_{2\times 2}$ contingency tables.
\textls[10]{
\bgroup
\begin{multline*}
\scriptsize
\begin{split}
C_{ORI_N,TAB_M} & = \begin{pmatrix}
  c_{ORI_1,TAB_1} & \hdots & c_{ORI_1,TAB_M} \\
  \vdots & \ddots & \vdots  \\
  c_{ORI_N,TAB_1} & & c_{ORI_N,TAB_M} 
 \end{pmatrix}
\end{split}
\normalsize
\end{multline*}
\vspace{-1em}
\scriptsize
\begin{multline*}
S_{ORI_N,TAB_M} = \sum^{}_{} C_{ORI_N,TAB_M} \cap \\(C_{ORI_i,TAB_j} \cup \sum^{}_{} ORI_i \cup \sum^{}_{} TAB_j)
\end{multline*}
\normalsize
\vspace{-1em}
\begin{multline*}
\scriptsize
\begin{split}
 C_{2\times 2} = 
 \begin{pmatrix}
  C_{ORI_i,TAB_j} & \sum^{}_{} ORI_i \\
  \sum^{}_{} TAB_j & S_{ORI_N,TAB_M}
 \end{pmatrix}
 \end{split}
 \normalsize
\end{multline*}
\egroup}

\clearpage
\newpage
\section{Preliminary Results}
\subsection{Online Data}
\begin{table}[H]
    \centering
    \begin{tabular}{c c}
        Song & Lyrics Word Cloud \\ \hline
        \begin{minipage}{1em}\vspace{-8em}\rotatebox[origin=c]{90}{\shortstack[c]{\scriptsize Justin Bieber\\ \scriptsize Love Yourself}}\end{minipage}  & \rotatebox[origin=c]{90}{\includegraphics[width=0.12\textwidth,clip,trim={4cm 2cm 3cm 6cm}]{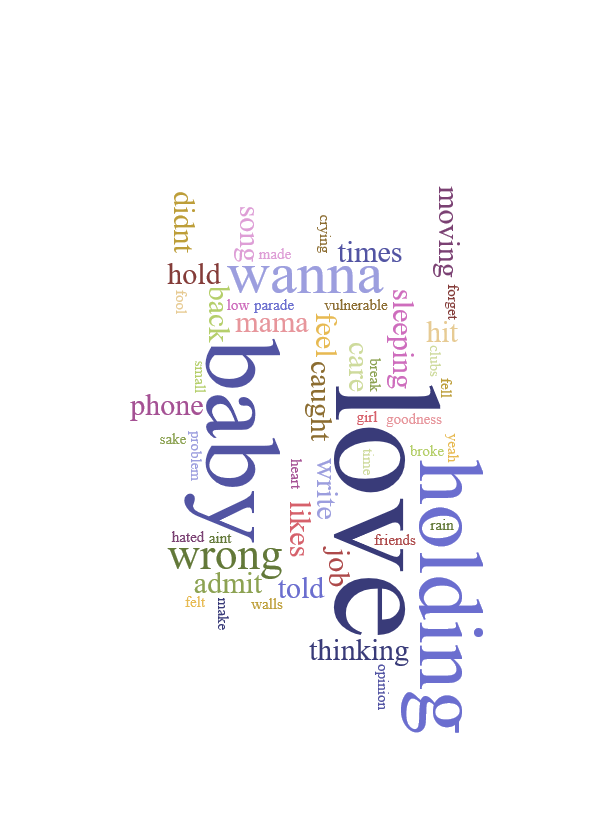}} \\ 
        \begin{minipage}{1em}\vspace{-8em}\rotatebox[origin=c]{90}{\shortstack[c]{\scriptsize Beyoncé\\ \scriptsize Halo}}\end{minipage} & \rotatebox[origin=c]{90}{\includegraphics[width=0.12\textwidth,clip,trim={5cm 7cm 4cm 6cm}]{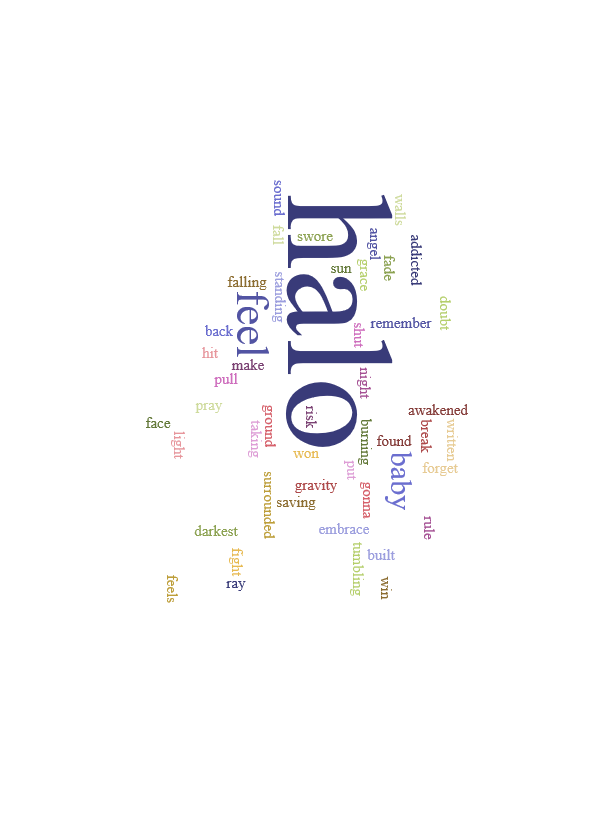}} \\ 
        \begin{minipage}{1em}\vspace{-8em}\rotatebox[origin=c]{90}{\shortstack[c]{\scriptsize Rihanna\\ \scriptsize Love On The Brain}}\end{minipage} & \rotatebox[origin=c]{90}{\includegraphics[width=0.12\textwidth,clip,trim={4cm 4cm 2cm 4cm}]{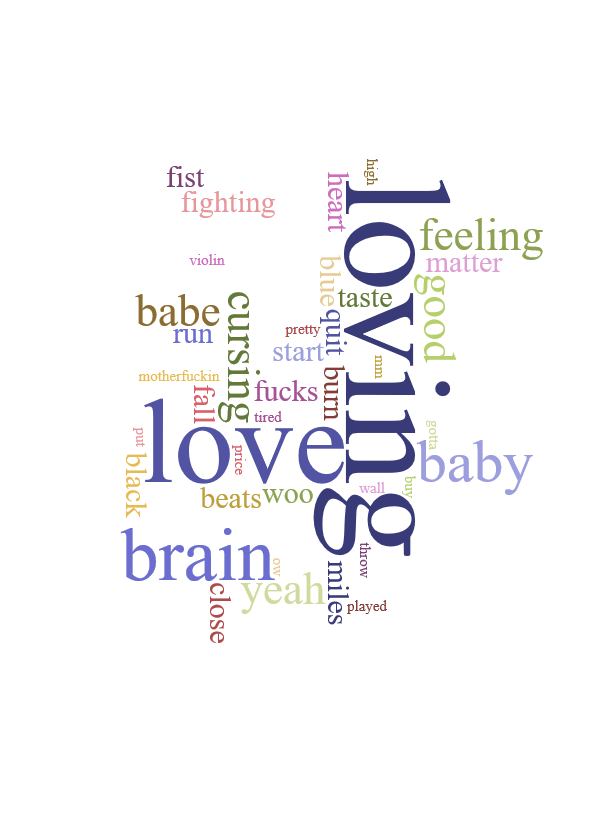}}\\
    \end{tabular}
    \vspace{-1.5em}
    \caption{Word cloud generated from the lyrics for the three English language songs}
    \label{tab:lyrics}
\end{table}
Table \ref{tab:lyrics} shows the word cloud for the three songs (`Love Yourself' by Justin Bieber, `Halo' by Beyoncé, and `Love On The Brain' by Rihanna) generated from the lyrics obtained online. The purpose of the word cloud is to give insight into which words are frequent in the lyrics. As it can be seen, the lyrics for all the songs often mention love and romantic feelings.

\begin{table}[H]
    \centering
\resizebox{0.9\columnwidth}{!}{%
    \begin{tabular}{c c c}
        Song & ASL Screenshot & Libras Screenshot \\ \hline
        \begin{minipage}{1em}\vspace{-4.5em}\rotatebox[origin=c]{90}{\shortstack[c]{\scriptsize Justin Bieber\\ \scriptsize Love Yourself}}\end{minipage} & \includegraphics[width=0.2\textwidth,clip,trim={0cm 1cm 0cm 1cm}]{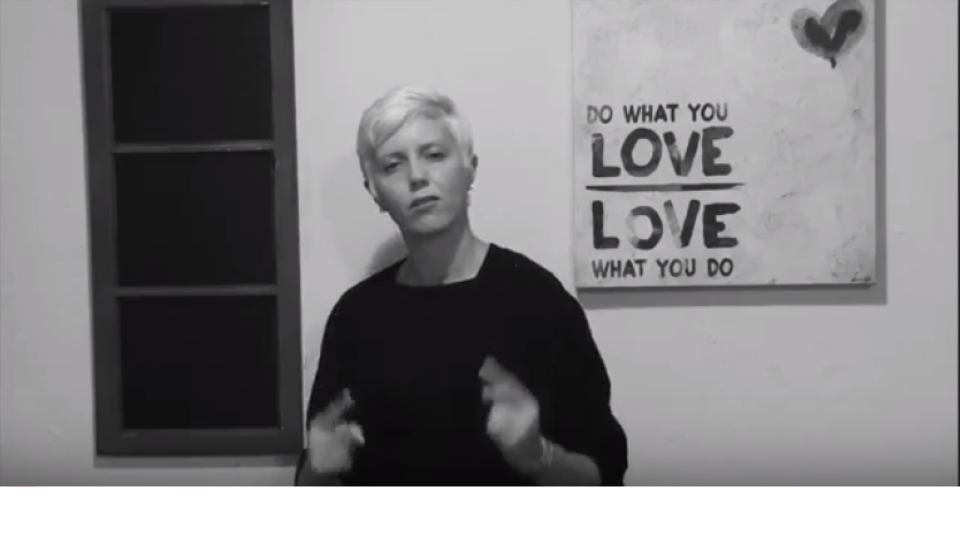} & \includegraphics[width=0.2\textwidth,clip,trim={0cm 1cm 0cm 0.85cm}]{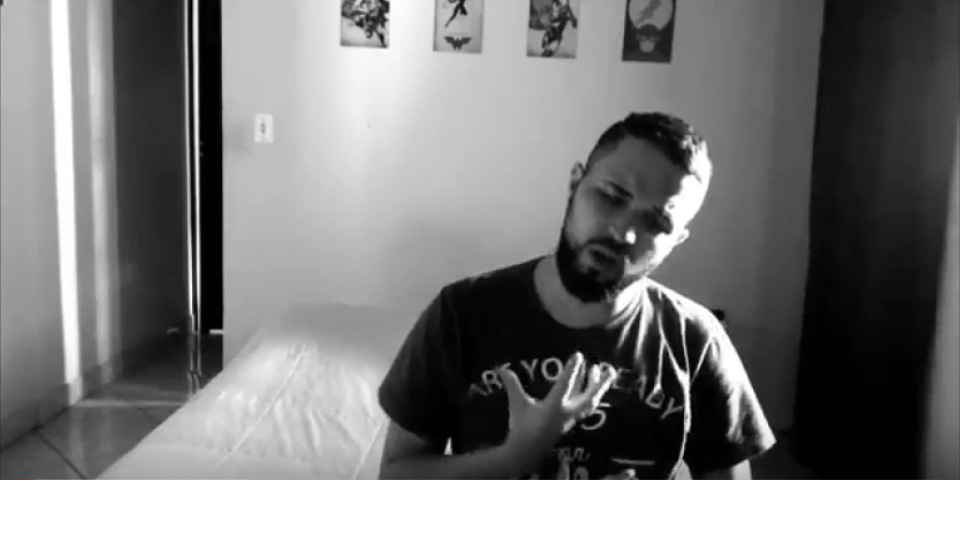} \\ 
        \begin{minipage}{1em}\vspace{-4.5em}\rotatebox[origin=c]{90}{\shortstack[c]{\scriptsize Beyoncé\\ \scriptsize Halo}}\end{minipage} & \includegraphics[width=0.2\textwidth,clip,trim={0cm 1cm 0cm 0cm}]{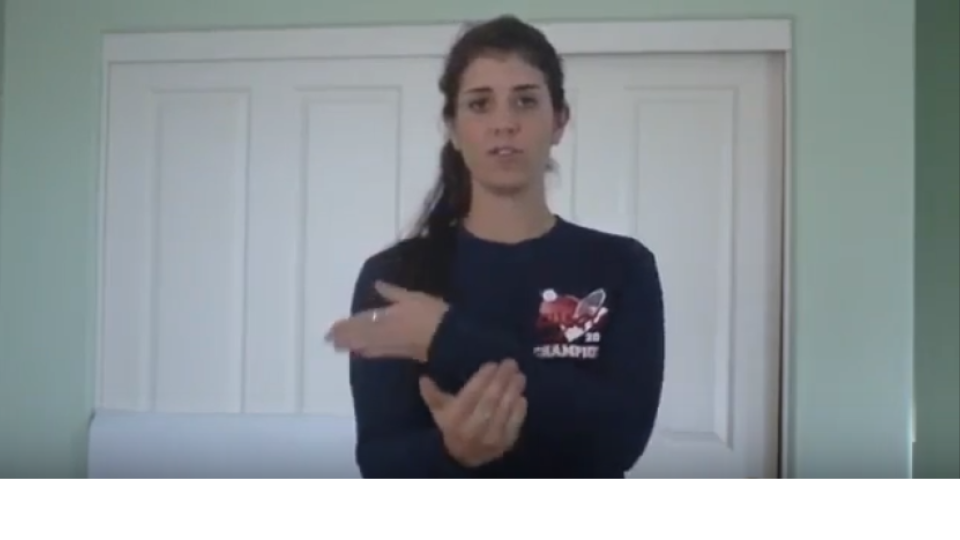} & \hspace{-1.3em}\rule{9.7em}{5.2em}\hspace{-9.7em}\includegraphics[width=0.17\textwidth,clip,trim={0cm 0cm 3cm 0cm}]{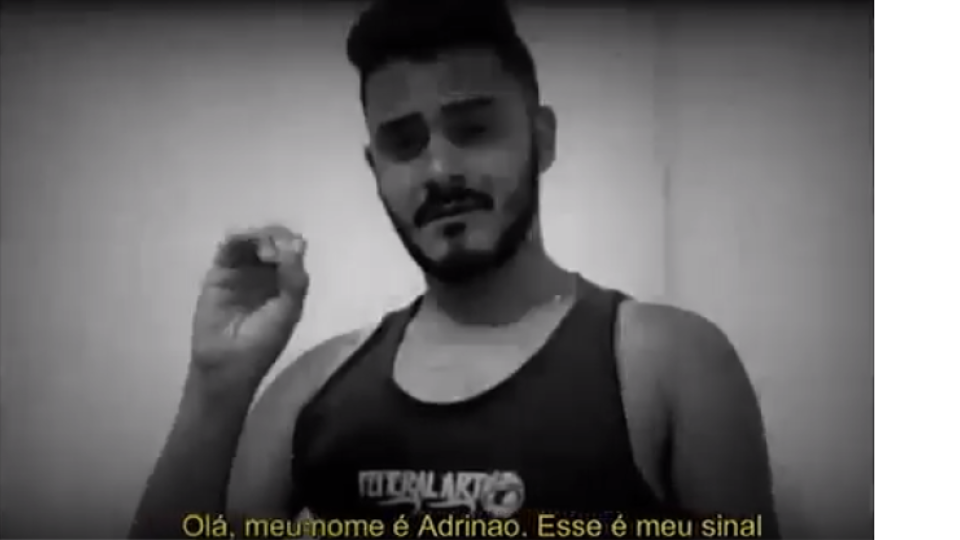} \\ 
        \begin{minipage}{1em}\vspace{-4.5em}\rotatebox[origin=c]{90}{\shortstack[c]{\scriptsize Rihanna\\ \scriptsize Love On The Brain}}\end{minipage} & \includegraphics[width=0.2\textwidth,clip,trim={0cm 1cm 0cm 0cm}]{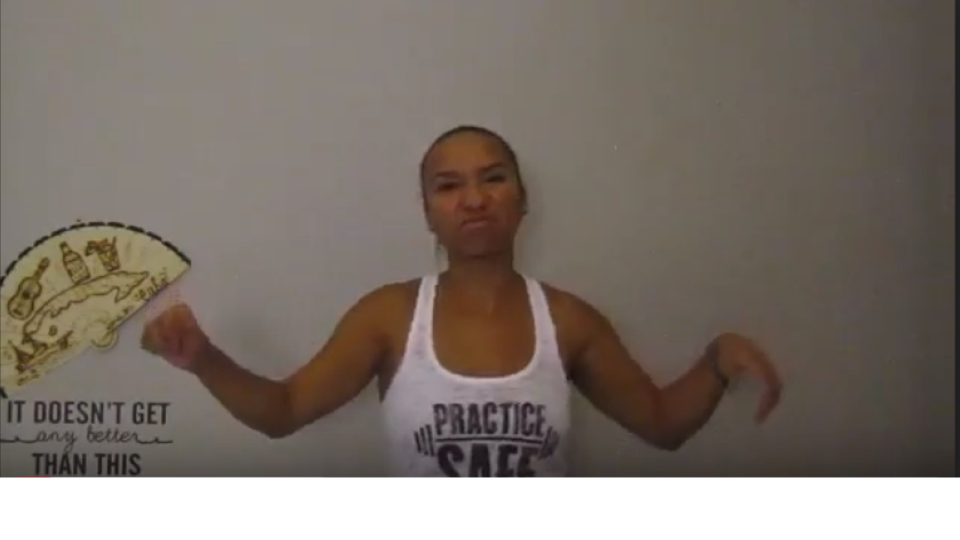} & \includegraphics[width=0.2\textwidth,clip,trim={0cm 1cm 0cm 0cm}]{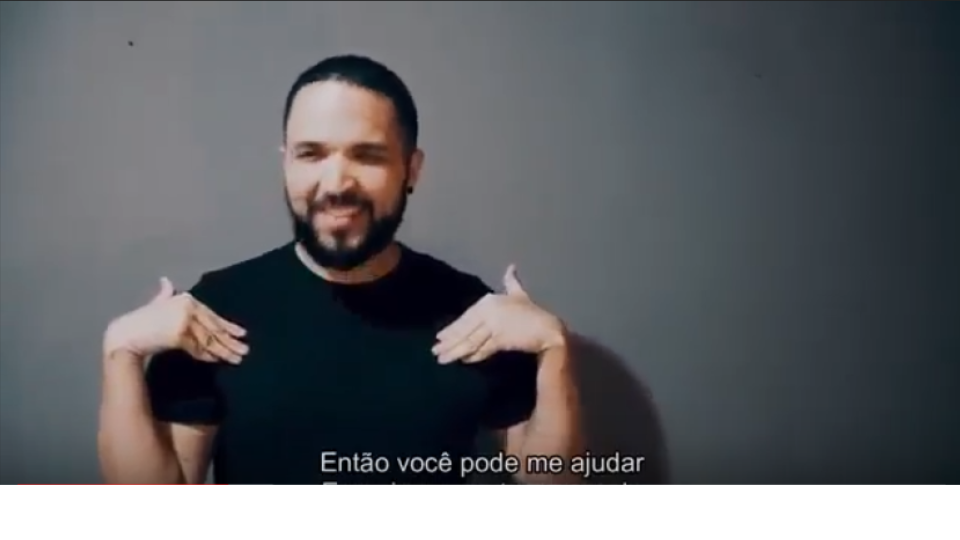} \\
    \end{tabular}
}
    \caption{Screenshots of the collected online data for three songs by three different artists with English spoken language interpreted by six different signers, three signers per sign language}
    \label{tab:online_data}
\end{table}
\placetextbox{0.38}{0.3315}{\crule[white]{3.6cm}{0.14cm}}%

Table \ref{tab:online_data} shows screenshots of the collected online data for three English songs performed by three different artists. The songs are interpreted in two sign languages by different signers. From the screenshots, it can be seen that the proximity of the signer to the camera varies. Some videos are edited by applying black and white or vintage camera filters. As a general rule, there is no camera movement, but the signers usually dance slightly to the songs.

\subsection{Data Filtering}
\vspace{-1em}
\begin{table}[H]
    \centering
\resizebox{0.9\columnwidth}{!}{%
    \begin{tabular}{c c c}
        Song & ASL Filtered & Libras Filtered \\ \hline
        \begin{minipage}{1em}\vspace{-4.5em}\rotatebox[origin=c]{90}{\shortstack[c]{\scriptsize Justin Bieber\\ \scriptsize Love Yourself}}\end{minipage} & \includegraphics[width=0.2\textwidth,clip,trim={0cm 1cm 0cm 1.1cm}]{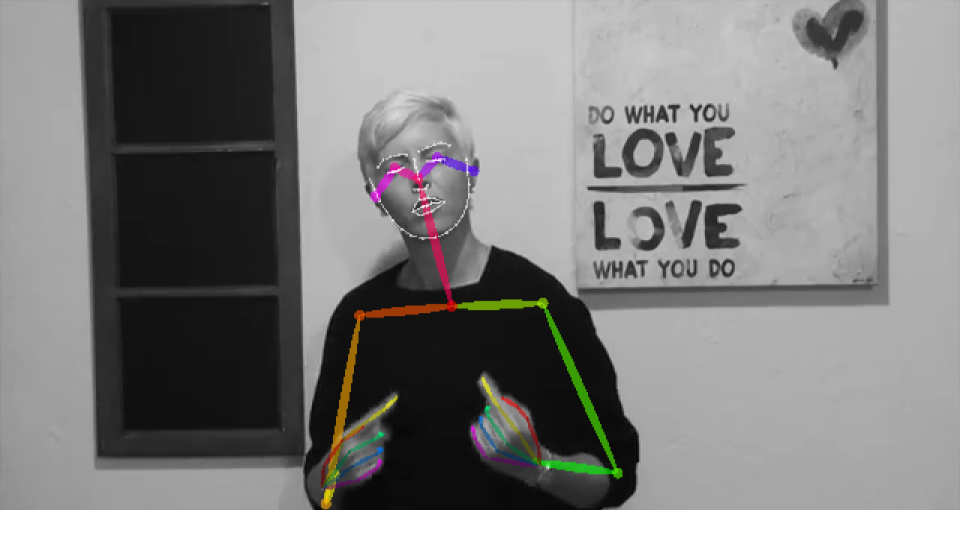} & \includegraphics[width=0.2\textwidth,clip,trim={0cm 1cm 0cm 0.4cm}]{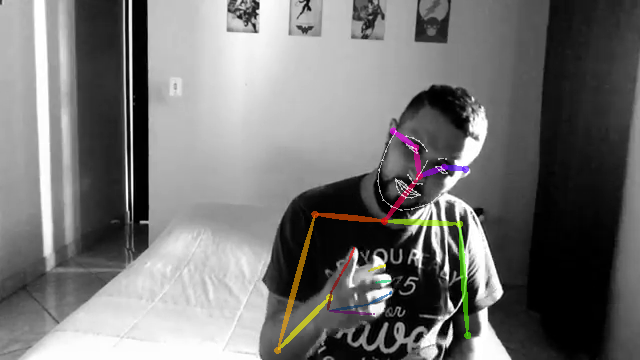} \\ 
        \begin{minipage}{1em}\vspace{-4.5em}\rotatebox[origin=c]{90}{\shortstack[c]{\scriptsize Beyoncé\\ \scriptsize Halo}}\end{minipage} & \includegraphics[width=0.2\textwidth,clip,trim={0cm 1cm 0cm 0cm}]{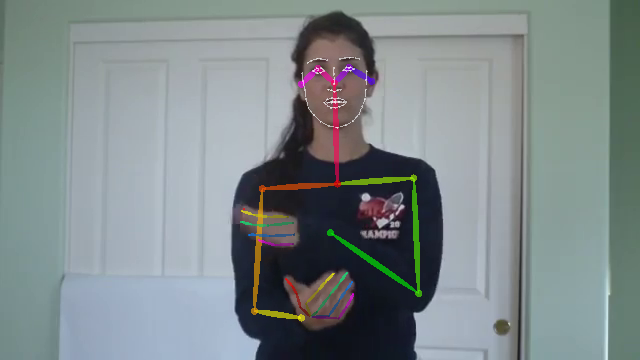} & \includegraphics[width=0.2\textwidth,clip,trim={0cm 1cm 0cm 0cm}]{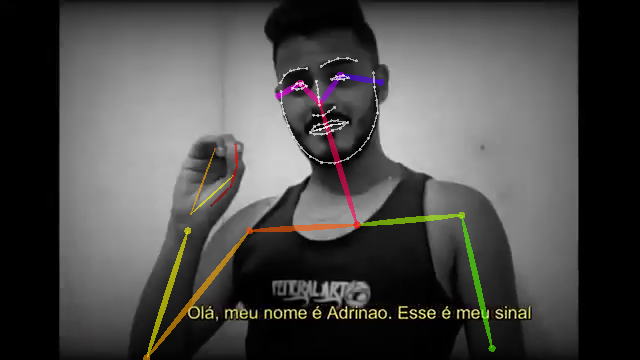} \\ 
        \begin{minipage}{1em}\vspace{-4.5em}\rotatebox[origin=c]{90}{\shortstack[c]{\scriptsize Rihanna\\ \scriptsize Love On The Brain}}\end{minipage} & \includegraphics[width=0.2\textwidth,clip,trim={0cm 1cm 0cm 0cm}]{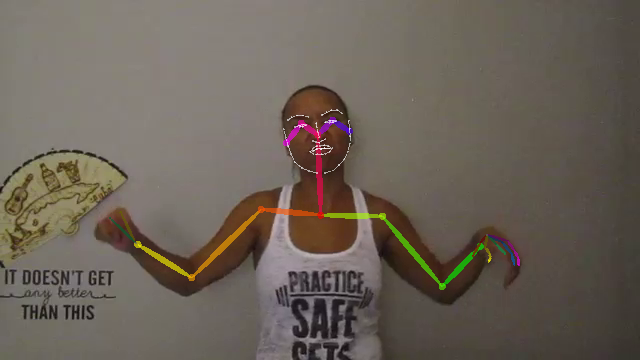} & \includegraphics[width=0.2\textwidth,clip,trim={0cm 1cm 0cm 0cm}]{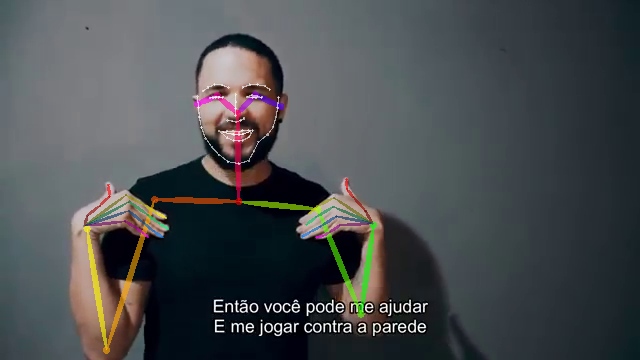} \\ 
    \end{tabular}
}
    \caption{Screenshots of the filtered online data after applying the OpenPose library. Visible skeletons on the screenshot means that the library was able to detect a human in the video and is tracking the pose, hand, and face key-points}
    \label{tab:res_filter}
\end{table}
Table \ref{tab:res_filter} shows screenshots of the filtered data after the OpenPose library has been applied to the data. The library was able to detect a human in the video and is tracking the pose, hand, and face key-points. Since we are not interested in keeping the integrity of the sequences of the frames, we simply discard the frames where key-points were not detected by the library.

\subsection{Discovering Patterns between Sign Languages}
\begin{table}[H]
    \centering
    \begin{tabular}{c c c}
        Song & ASL HamNoSys & Libras HamNoSys \\ \hline
        \begin{minipage}{1em}\vspace{-6em}\rotatebox[origin=c]{90}{\shortstack[c]{\scriptsize Justin Bieber\\ \scriptsize Love Yourself}}\end{minipage} & \includegraphics[width=0.2\textwidth,clip,trim={0cm 0cm 0cm 0cm}]{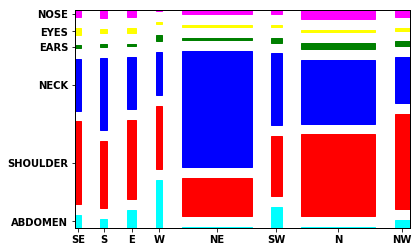} & \includegraphics[width=0.2\textwidth,clip,trim={0cm 0cm 0cm 0cm}]{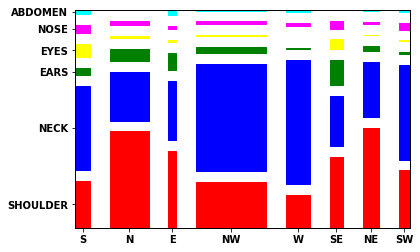} \\ 
        \begin{minipage}{1em}\vspace{-6em}\rotatebox[origin=c]{90}{\shortstack[c]{\scriptsize Beyoncé\\ \scriptsize Halo}}\end{minipage} & \includegraphics[width=0.2\textwidth,clip,trim={0cm 0cm 0cm 0cm}]{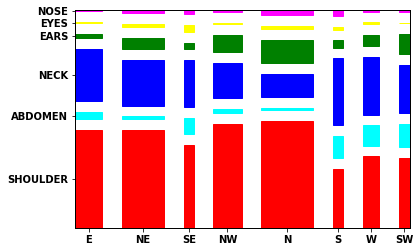} & \includegraphics[width=0.2\textwidth,clip,trim={0cm 0cm 0cm 0cm}]{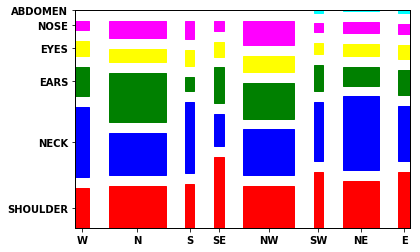} \\ 
        \begin{minipage}{1em}\vspace{-5.5em}\rotatebox[origin=c]{90}{\shortstack[c]{\scriptsize Rihanna\\ \scriptsize Love On The Brain}}\end{minipage} & \includegraphics[width=0.2\textwidth,clip,trim={0cm 0cm 0cm 0cm}]{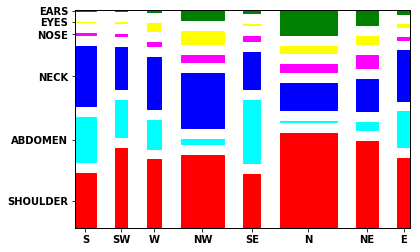} & \includegraphics[width=0.2\textwidth,clip,trim={0cm 0cm 0cm 0cm}]{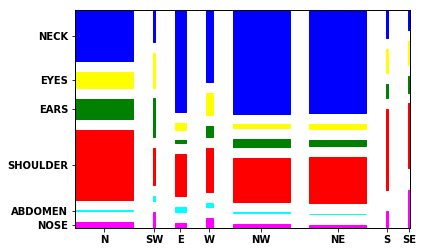} \\
    \end{tabular}
    \caption{Location/orientation relative frequencies for each video for each sign language}
    \label{tab:res_patterns}
\end{table}

Table \ref{tab:res_patterns} shows the relative frequencies of the location/orientation combinations for each video and each sign language. We can observe that the Libras, on one hand, has less abdomen activity than the ASL (indicated in light blue) while, on the other hand, Libras has more neck and ears activity than the ASL (dark blue and green respectively). Both sign languages have more pointing up direction of the hands as opposed to other possible directions (wider NE/N/NW columns).

\subsection{Data Analysis / Modelling}
\vspace{-1.3em}
\begin{minipage}{.2\textwidth}
\begin{figure}[H]
    \centering
    \includegraphics[width=1.1\textwidth, clip,trim={6cm 6cm 5cm 11cm}]{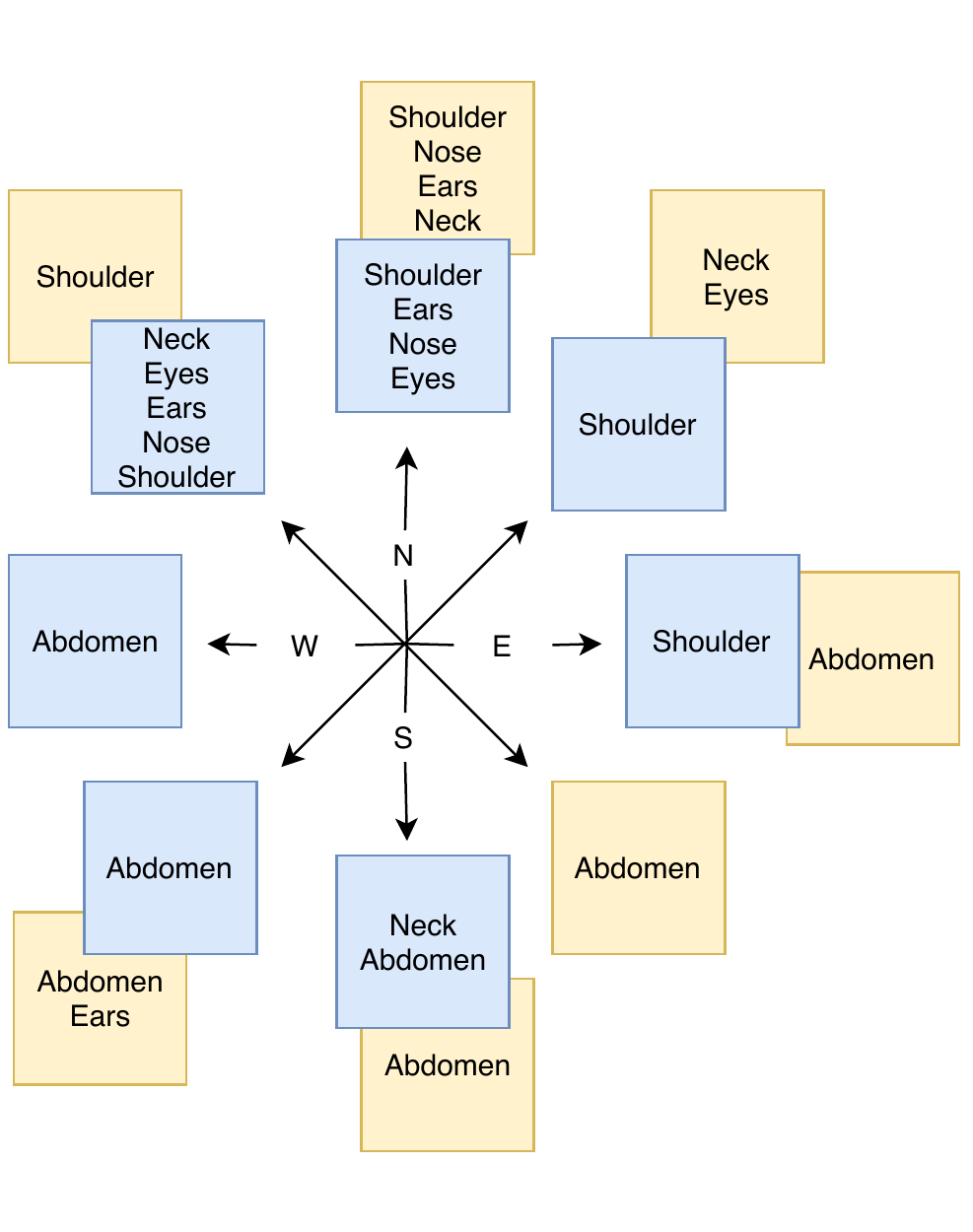}
    \caption*{a) ASL}
\end{figure}
\end{minipage}
\hspace{1em}
\begin{minipage}{.2\textwidth}
\begin{figure}[H]
    \centering
    \includegraphics[width=1.1\textwidth, clip,trim={5.5cm 6.5cm 5cm 11cm}]{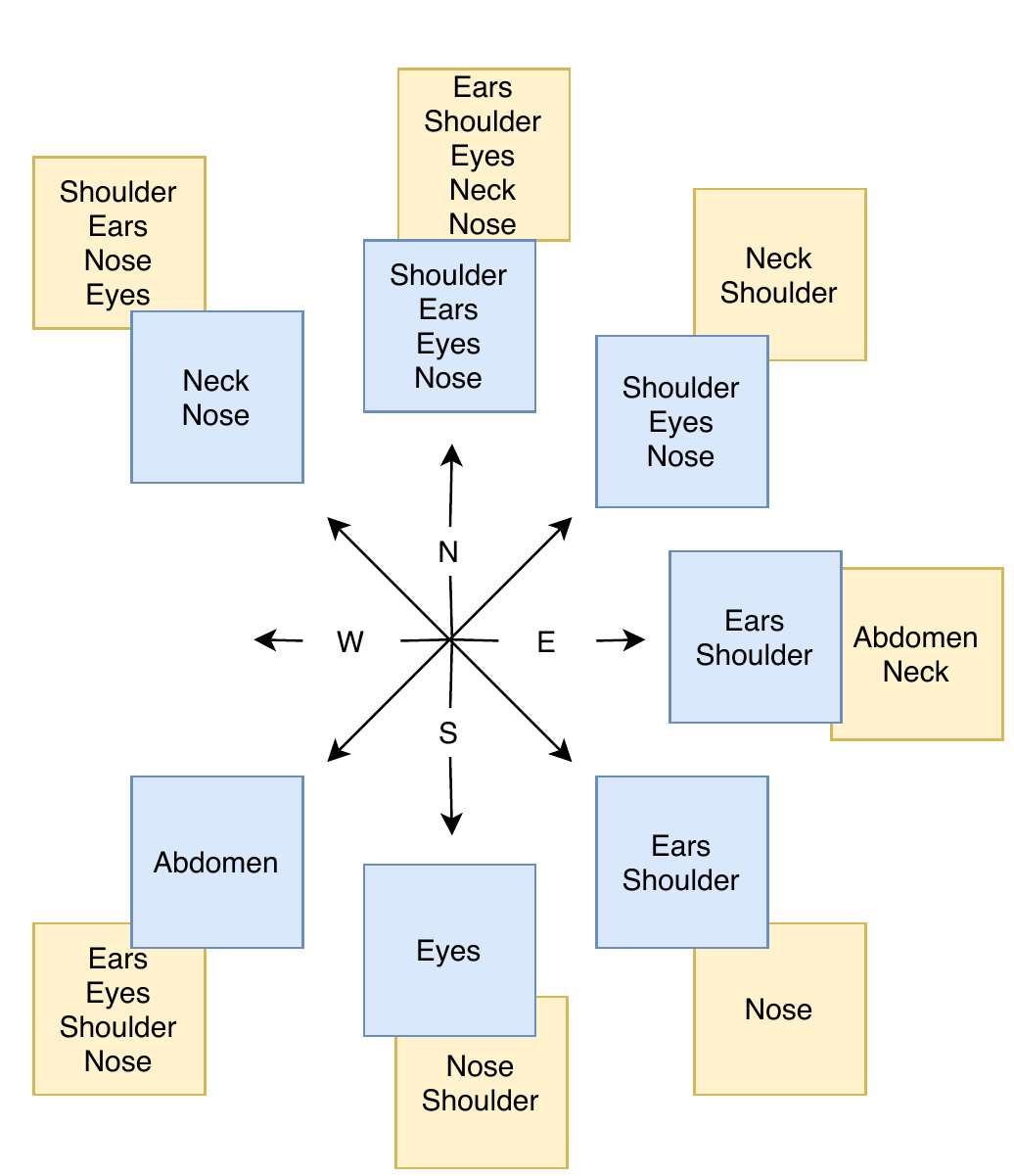}
    \vspace{0.15em}
    \caption*{b) Libras}
\end{figure}
\end{minipage}
\begin{figure}[H]
    \centering
    \includegraphics[width=0.2\textwidth, clip,trim={6.1cm 4cm 5.5cm 24cm}]{pics/significant_ORI_TAB_correlation-all_libras}
    \label{fig:asl_codep}
\end{figure}
\begin{figure}[H]
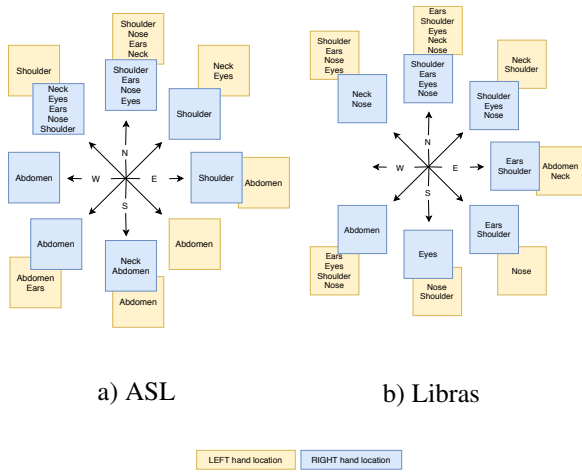

    \centering
    \caption{Significant location and orientation phonological parameter co-dependences in a) ASL and b) Libras with Bonferroni-adjusted Chi-squared p-value $<$ 0.001 for both sign languages}
    \label{fig:asl_codep}
\end{figure}

Having visually analysed the frequencies of the location/orientation combinations, we are interested in finding the significant combinations for each hand that prevails in the collected data and compare the two sign languages based on this analysis.

\subsubsection{Phonological Parameter Co-Dependence\\}
Figure \ref{fig:asl_codep} shows significant location/orientation co-dependences for each sign language after the Bonferroni-adjusted Chi-squared p-value analysis. We can see that both hands tend to point up at the upper side of the body, which is similar for the both sign languages. Libras, however, has more activity with both hands at the upper part of the body than the ASL. As a matter of fact, Libras has more activity with both hands around all the parts of the body. In ASL, on the other hand, the left hand is less mobile than the right hand. This could be explained by the fact that the signers in Libras were left-handed, but we do not have this information available to verify this speculation.
Some significant co-dependences are unusual, for example, pointing down at the upper body level, which may feel unnatural and slightly contradicts the past findings by \newcite{cooper2012sign} stating that a subset of the `comfortable' hand configurations are assumed more often during the signing, independent of the sign language. This can also be explained by the fact that the signers in the video are slightly dancing to the music, which may affect the signing orientation.

It is worth mentioning that the co-dependence analysis results of the two languages may change with the data. For example, if songs with a different sentiment were taken for the analysis. More data is needed to experiment this further.


\section{Conclusion}
In this work, we have showed the preliminary results of mining sign language data acquired from the internet for automated data-driven sign language processing. We have created a pipeline that downloads the videos of the interpreted songs from the internet, applies filtering of the data and then finds patterns in the data based on the HamNoSys notation that is often used for the annotation of the sign languages. This method could also be used for querying videos in large datasets. Finally, we compare two historically different sign languages (ASL and Libras) by their location/orientation co-dependencies present in the collected data and show that, despite there being little historical background of the two languages interacting, they still share similar signing patterns with small variations in the flexibility of the hands, which can be explained by the fact that people converge to the usage of the `comfortable' hand configurations.

Future work will compare even more historically unrelated sign languages and look at the interpretations of a greater number of songs, in order to have a more accurate comparison of the signing patterns across the sign languages.

\section{Acknowledgements}
This work was supported by the Heriot-Watt University School of Engineering \& Physical Sciences James Watt Scholarship and Engineering and Physical Sciences Research Council (EPSRC), as part of the CDT in Robotics and Autonomous Systems at Heriot-Watt University and The University of Edinburgh (Grant reference EP/L016834/1)

\section{Bibliographical References}\label{reference}

\bibliographystyle{lrec}
\bibliography{lrec2020W-xample-kc}

\begin{thebibliography}{}

\bibitem[\protect\citename{Belissen}2018]{belissen2019sign}
Belissen, V.
\newblock (2018).
\newblock Sign language video analysis for automatic recognition and detection.
\newblock In {\em Proceedings of the 20th International ACM SIGACCESS
  Conference on Computers and Accessibility}.

\bibitem[\protect\citename{Bilge \bgroup et al.\egroup }2019]{bilge2019zero}
Bilge, Y.~C., Ikizler-Cinbis, N., and Cinbis, R.~G.
\newblock (2019).
\newblock Zero-shot sign language recognition: Can textual data uncover sign
  languages?
\newblock In {\em Proceedings of the British Machine Vision Conference}.

\bibitem[\protect\citename{Bland and Altman}1995]{Bland170}
Bland, J.~M. and Altman, D.~G.
\newblock (1995).
\newblock Multiple significance tests: the bonferroni method.
\newblock {\em BMJ}, 310(6973):170.

\bibitem[\protect\citename{Buehler \bgroup et al.\egroup
  }2009]{buehler2009learning}
Buehler, P., Zisserman, A., and Everingham, M.
\newblock (2009).
\newblock Learning sign language by watching tv (using weakly aligned
  subtitles).
\newblock In {\em Proceedings of the IEEE conference on Computer Vision and
  Pattern Recognition}.

\bibitem[\protect\citename{Buehler \bgroup et al.\egroup
  }2010]{buehler2010employing}
Buehler, P., Everingham, M., and Zisserman, A.
\newblock (2010).
\newblock Employing signed tv broadcasts for automated learning of {British
  Sign Language}.
\newblock In {\em Proceedings of the 4th Workshop on the Representation and
  Processing of Sign Languages}.

\bibitem[\protect\citename{Camg\"oz \bgroup et al.\egroup
  }2018]{cihan2018neural}
Camg\"oz, N.~C., Hadfield, S., Koller, O., Ney, H., and Bowden, R.
\newblock (2018).
\newblock Neural sign language translation.
\newblock In {\em Proceedings of the Conference on Computer Vision and Pattern
  Recognition}.

\bibitem[\protect\citename{Cao \bgroup et al.\egroup }2017]{cao2017realtime}
Cao, Z., Simon, T., Wei, S.-E., and Sheikh, Y.
\newblock (2017).
\newblock Realtime multi-person {2D} pose estimation using part affinity
  fields.
\newblock In {\em Proceedings of the IEEE conference on Computer Vision and
  Pattern Recognition}.

\bibitem[\protect\citename{Cao \bgroup et al.\egroup }2018]{cao2018openpose}
Cao, Z., Hidalgo, G., Simon, T., Wei, S.-E., and Sheikh, Y.
\newblock (2018).
\newblock Open{P}ose: realtime multi-person 2{D} pose estimation using {P}art
  {A}ffinity {F}ields.
\newblock In {\em arXiv preprint arXiv:1812.08008}.

\bibitem[\protect\citename{Cooper and Bowden}2007]{cooper2007large}
Cooper, H. and Bowden, R.
\newblock (2007).
\newblock Large lexicon detection of sign language.
\newblock In {\em Proceedings of the International Workshop on Human-Computer
  Interaction}.

\bibitem[\protect\citename{Cooper \bgroup et al.\egroup }2012]{cooper2012sign}
Cooper, H., Ong, E.-J., Pugeault, N., and Bowden, R.
\newblock (2012).
\newblock Sign language recognition using sub-units.
\newblock {\em Journal of Machine Learning Research}, 13(Jul):2205--2231.

\bibitem[\protect\citename{Desblache}2019]{Desblache2019}
Desblache, L., (2019).
\newblock {\em How is Music Translated? Mapping the Landscape of Music
  Translation}, pages 219--264.
\newblock Palgrave Macmillan UK, London.

\bibitem[\protect\citename{Hanke}2004]{hanke2004hamnosys}
Hanke, T.
\newblock (2004).
\newblock Hamnosys-representing sign language data in language resources and
  language processing contexts.
\newblock In {\em Proceedings of the Workshop on Representation and processing
  of sign languages (LREC 2004)}.

\bibitem[\protect\citename{Joze and Koller}2018]{joze2018ms}
Joze, H. R.~V. and Koller, O.
\newblock (2018).
\newblock {MS-ASL}: A large-scale data set and benchmark for understanding
  {American Sign Language}.
\newblock In {\em Proceedings of the British Machine Vision Conference}.

\bibitem[\protect\citename{Ko \bgroup et al.\egroup }2019]{ko2019neural}
Ko, S.-K., Kim, C.~J., Jung, H., and Cho, C.
\newblock (2019).
\newblock Neural sign language translation based on human keypoint estimation.
\newblock {\em Applied Sciences}, 9(13):2683.

\bibitem[\protect\citename{{Koller} \bgroup et al.\egroup }2016]{7780781}
{Koller}, O., {Ney}, H., and {Bowden}, R.
\newblock (2016).
\newblock Deep hand: How to train a cnn on 1 million hand images when your data
  is continuous and weakly labelled.
\newblock In {\em Proceedings of the IEEE conference on Computer Vision and
  Pattern Recognition}.

\bibitem[\protect\citename{Konrad}2015]{reinerkonrad2015}
Konrad, R.
\newblock (2015).
\newblock {DGS} corpus annotation guidelines.
\newblock In {\em Proceedings of Digging into Signs Workshop: Developing
  Annotation Standards for Sign Language Corpora}.

\bibitem[\protect\citename{Mocialov \bgroup et al.\egroup
  }2017]{Mocialov2017TowardsCS}
Mocialov, B., Turner, G., Lohan, K.~S., and Hastie, H.
\newblock (2017).
\newblock Towards continuous sign language recognition with deep learning.
\newblock In {\em Proceedings of Workshop of Creating Meaning With Robot
  Assistants (Humanoids 2017)}.

\bibitem[\protect\citename{Nandy \bgroup et al.\egroup
  }2010]{10.1007/978-3-642-12214-9_18}
Nandy, A., Prasad, J.~S., Mondal, S., Chakraborty, P., Nandi, G.~C.", e. V.~V.,
  Vijayakumar, R., Debnath, N.~C., Stephen, J., Meghanathan, N.,
  Sankaranarayanan, S., Thankachan, P.~M., Gaol, F.~L., and Thankachan, N.
\newblock (2010).
\newblock {Recognition of Isolated Indian Sign Language Gesture in Real Time}.
\newblock In {\em Proceedings of the Conference on Information Processing and
  Management}.

\bibitem[\protect\citename{{\"O}stling \bgroup et al.\egroup
  }2018]{ostling2018visual}
{\"O}stling, R., B{\"o}rstell, C., and Courtaux, S.
\newblock (2018).
\newblock Visual iconicity across sign languages: Large-scale automated video
  analysis of iconic articulators and locations.
\newblock {\em Frontiers in psychology}, 9:725.

\bibitem[\protect\citename{Simon \bgroup et al.\egroup }2017]{simon2017hand}
Simon, T., Joo, H., Matthews, I., and Sheikh, Y.
\newblock (2017).
\newblock Hand keypoint detection in single images using multiview
  bootstrapping.
\newblock In {\em Proceedings of the IEEE conference on Computer Vision and
  Pattern Recognition}.

\bibitem[\protect\citename{Simonyan and Zisserman}2015]{simonyan2014deep}
Simonyan, K. and Zisserman, A.
\newblock (2015).
\newblock Very deep convolutional networks for large-scale image recognition.
\newblock In {\em Proceedings of the International Conference on Learning
  Representations}.

\bibitem[\protect\citename{Stefanov and Beskow}2017]{stefanov2017real}
Stefanov, K. and Beskow, J.
\newblock (2017).
\newblock {A Real-Time Gesture Recognition System for Isolated Swedish Sign
  Language Signs}.
\newblock In {\em Proceedings of the 4th European and 7th Nordic Symposium on
  Multimodal Communication (MMSYM)}.

\bibitem[\protect\citename{Stoll \bgroup et al.\egroup
  }2019]{stoll2019text2sign}
Stoll, S., Camg\"oz, N.~C., Hadfield, S., and Bowden, R.
\newblock (2019).
\newblock Text2sign: Towards sign language production using neural machine
  translation and generative adversarial networks.
\newblock {\em International Journal of Computer Vision}, pages 1--18.

\bibitem[\protect\citename{Takayama and Takahashi}2018]{takayama2018sign}
Takayama, N. and Takahashi, H.
\newblock (2018).
\newblock Sign words annotation assistance using {Japanese Sign Language} words
  recognition.
\newblock In {\em Proceedings of the International Conference on Cyberworlds
  (CW)}.

\bibitem[\protect\citename{Wei \bgroup et al.\egroup }2016]{wei2016cpm}
Wei, S.-E., Ramakrishna, V., Kanade, T., and Sheikh, Y.
\newblock (2016).
\newblock Convolutional pose machines.
\newblock In {\em Proceedings of the IEEE conference on Computer Vision and
  Pattern Recognition}.

\bibitem[\protect\citename{Yuan \bgroup et al.\egroup }2019]{yuan2019large}
Yuan, T., Sah, S., Ananthanarayana, T., Zhang, C., Bhat, A., Gandhi, S., and
  Ptucha, R.
\newblock (2019).
\newblock Large scale sign language interpretation.
\newblock In {\em Proceedings of the 14th IEEE International Conference on
  Automatic Face \& Gesture Recognition (FG 2019)}, pages 1--5. IEEE.

\bibitem[\protect\citename{{Zafrulla} \bgroup et al.\egroup }2011]{5771325}
{Zafrulla}, Z., {Brashear}, H., {Presti}, P., {Hamilton}, H., and {Starner}, T.
\newblock (2011).
\newblock {CopyCat: An American Sign Language game for deaf children}.
\newblock In {\em Proceedings of the International Conference on Automatic Face
  and Gesture Recognition}, pages 647--647, March.

\bibitem[\protect\citename{Zhou \bgroup et al.\egroup }2009]{zhou2009adaptive}
Zhou, Y., Chen, X., Zhao, D., Yao, H., and Gao, W.
\newblock (2009).
\newblock Adaptive sign language recognition with exemplar extraction and
  map/ivfs.
\newblock {\em IEEE signal processing letters}, 17(3):297--300.

\end{thebibliography}


\end{document}